\documentclass[letterpaper, 10 pt, conference]{ieeeconf}  

\IEEEoverridecommandlockouts                              
\IEEEoverridecommandlockouts

\usepackage{flushend}

\usepackage{url}

\usepackage{amsmath}
\usepackage{graphics}
\usepackage{graphicx}
\usepackage{float}
\usepackage{amsmath,amssymb}
\DeclareMathOperator{\E}{\mathbb{E}}

\usepackage{todonotes}

\usepackage{caption}
\makeatletter
\let\NAT@parse\undefined
\makeatother
\usepackage[numbers]{natbib}

\title{\LARGE \bf ClusterNet: 3D Instance Segmentation in RGB-D Images}

\author{Lin Shao, Ye Tian, and Jeannette Bohg
\thanks{The authors are with the Computer Science Department of Stanford University, CA, USA.
        {\tt\small [lins2,yetian1,bohg]@stanford.edu}}%
}

\begin{document}

\maketitle
\begin{abstract}
We propose a method for instance-level segmentation that uses RGB-D data as input and provides detailed information about the location, geometry and number of {\em individual\/} objects in the scene. This level of understanding is fundamental for autonomous robots. It enables safe and robust decision-making under the large uncertainty of the real-world. In our model, we propose to use the first and second order moments of the object occupancy function to represent an object instance. We train an hourglass Deep Neural Network (DNN) where each pixel in the output votes for the 3D position of the corresponding object center and for the object's size and pose. The final instance segmentation is achieved through clustering in the space of moments. The object-centric training loss is defined on the output of the clustering. Our method outperforms the state-of-the-art instance segmentation method on our synthesized dataset. We show that our method generalizes well on real-world data achieving visually better segmentation results.
\end{abstract}

\section{Introduction}
Instance segmentation is a notoriously hard problem in the diverse and challenging conditions of real-world environments. Yet, it is a fundamental ability for autonomous robots as it enables safe and robust decision-making under the large uncertainty of the real-world. Autonomous robots need to detect, outline and track animate and inanimate objects in real-time to decide how to act next. Unlike category-level segmentation, instance segmentation aims to provide detailed information about the location, geometry and number of {\em individual\/} objects. Instance segmentation is particularly challenging in scenarios where objects are heavily overlapping with each other. Furthermore, an image can contain an arbitrary number of object instances. The labeling of these instances has to be permutation-invariant, i.e. it does not matter which specific label an instance receives, as long as the label is different from all other object instance labels.  

\begin{figure}
  \centering
    \centering
    \includegraphics[width=1.0\columnwidth]{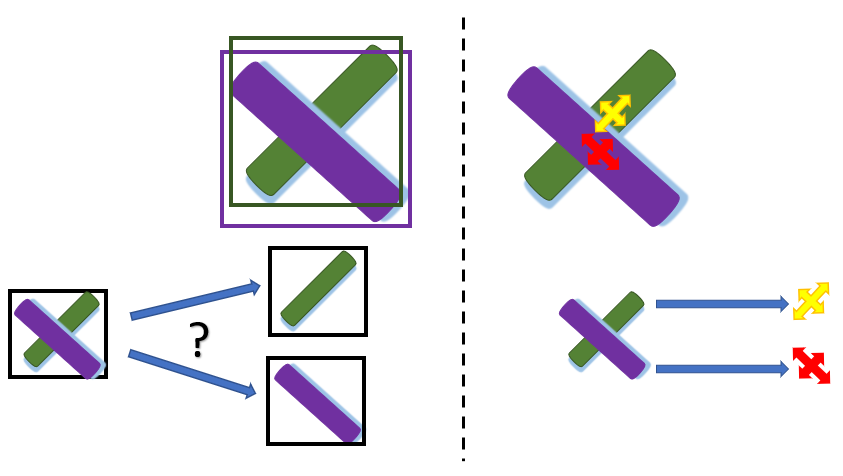}
    \caption{\label{fig:occlusion}Challenging scene for segmentation methods based on bounding boxes. Left: Two bounding boxes overlap. Due to ambiguity, methods like Mask R-CNN \cite{he2017mask} are challenged to segment objects within the proposed boxes. Right: Two overlapping objects have different features (3D object position, pose and size) in our proposed feature space. They can therefore be well seperated. Our model makes per-pixel predictions of these features.} 
 \end{figure}

Current state of the art approaches mainly operate on single RGB images. They had enormous success by leveraging large-scale, annotated datasets \citep{lin2014microsoft} and large-capacity, deep neural networks \citep{he2017mask}. The most successful approaches such as Mask R-CNN \citep{he2017mask} rely on a region proposal process. A first module generates object proposals in the form of 2D bounding boxes. These bounding boxes serve as input to a module performing object recognition and segmentation within these boxes. Such networks are challenged in cluttered scenarios with heavily overlapping objects where the region proposal may already contain multiple instances (for an example see Fig.~\ref{fig:occlusion}).  

Inspired by the benefits of including depth data in robot perception tasks~\citep{gupta2014learning}~\citep{han2013enhanced}, we propose a novel instance segmentation method that avoids region proposals and therefore the aforementioned challenges. In our approach, the receptive field of each prediction is not constrained to the detection box of the corresponding instance. In this way, it can utilize more global information than region-proposal based method that are restricted to the inside of the proposed region.
Central to our approach is {\em (i) to exploit depth data and 
(ii) an explicit embedding in a feature space related to 3D object properties.\/} This model is not merely a variation of Mask R-CNN~\cite{he2017mask} but rethinks how depth data can be best exploited. 


Specifically, we use the first and second order moments of the object occupancy function to represent an object instance. It enables us to transform the problem of instance-level segmentation into a general regression problem. We train an hourglass Deep Neural Network (DNN) where each pixel in the output votes for the relative position of the corresponding object center (first order moment) and for the object's size and pose (second order moment). The final instance segmentation is achieved through clustering in the space of moments. The carefully-designed, object-centric training loss is defined on the output of the clustering. 
We show that the combination of the proposal-free method with depth data leads to improved accuracy and robustness of instance segmentation under challenging real-world conditions. Furthermore, the proposed explicit embedding could be more readily used in a robot manipulation scenario.

Our primary contributions are: (1) proposing a novel deep learning framework for 3D instance segmentation based on RGB-D images. (2) providing an extensive quantitative evaluation that shows state-of-the-art performance on a difficult, synthetic dataset targeted at robotic manipulation and validate our design choices as well as qualitative results on real data for understanding the strengths and limitations of our method.\label{sec:intro}

\section{Related Work}\label{sec:related}
Instance segmentation methods can be divided into two main approaches: methods based on region proposals and proposal-free methods. 
Approaches based on region proposals first generate these proposals (e.g. in the form of bounding boxes). Then in these regions, objects are detected and segmented. Proposal-free approaches drop the proposal process. They either sequentially segment object instances or use different object representations. In the following, we review approaches in each of the two categories.

\subsection{Region Proposal Based Approaches}
\citet{hariharan2014simultaneous} propose one of the first works that address instance-level segmentation and uses the region proposal pipeline. It employs a convolutional neural network to generate bounding boxes which are then classified into object categories. Tightness of the boxes is improved bounding box regression. A class-specific segmentation is then performed in this bounding box to simultaneously detect and segment the object. \citet{he2017mask} use Fast-RCNN boxes \cite{he2016deep} and build a multi-stage pipeline to extract features, classify and segment the object. \citet{gupta2014learning} performs instance segmentation based on region proposals in RGB-D images. The authors use decision forests to predict a foreground mask in a detection bounding box.

Approaches based on region proposals achieve state-of-the-art results on large-scale datasets such as for example COCO \citep{lin2014microsoft} and Cityscapes \citep{cordts2016cityscapes}. Such approaches are however challenged when facing small objects or objects that are strongly occluding each other. In these cases, a bounding box may contain multiple objects and it becomes difficult to detect and segment the correct one as visualized in Fig. \ref{fig:occlusion}). However, occlusions are very common especially in robotic manipulation with scenarios such as bin-picking or when cleaning a cluttered counter top.

\subsection{Region Proposal Free Approaches}
Proposal-free approaches differ from each other mainly in the employed object representation. \citet{de2017semantic} train a neural network learning pixel features of instances in an unsupervised fashion based on a discriminative loss \cite{schroff2015facenet}. The loss function encourages the network to map each pixel to a point in feature space so that pixels belonging to the same instance are close to each other while different instances are separated by a wide margin. \citet{newell2017associative} propose an approach that
teaches a network to simultaneously output detections and
group assignments through implicit embedding. Different from these implicit embedding methods, our work leverages the explicit embedding which is built upon object geometry that we qualitatively show to have a small transfer gap to real data. Furthermore, the attributes (object center and object covariance matrix encoding pose and shape) are useful features for robotic manipulation. 

\citet{liang2017proposal} predict pixel-wise feature vectors representing the ground truth bounding box of the instance it belongs to. Leveraging another sub-network that predicts an object count, they cluster the output of the network into individual instances using spectral clustering \cite{ng2002spectral}. However predicting the exact number of instances is difficult and a wrongly predicted number harms the segmentation result. We propose to use object features such as its 3D centroid and pose that are predicted per pixel. These form the input to the clustering method to infer the final number of instances.

\section{Technical Approach}\label{sec:method}
We are proposing a 3D region-proposal-free approach towards instance segmentation. Specifically, we propose a new object representation that is predicted per pixel. This is followed by a clustering steps in this new feature space. The training loss is defined on the output of the clustering.
\subsection{Object Representation}
\begin{figure}
    \centering
    \includegraphics[width=0.9\columnwidth]{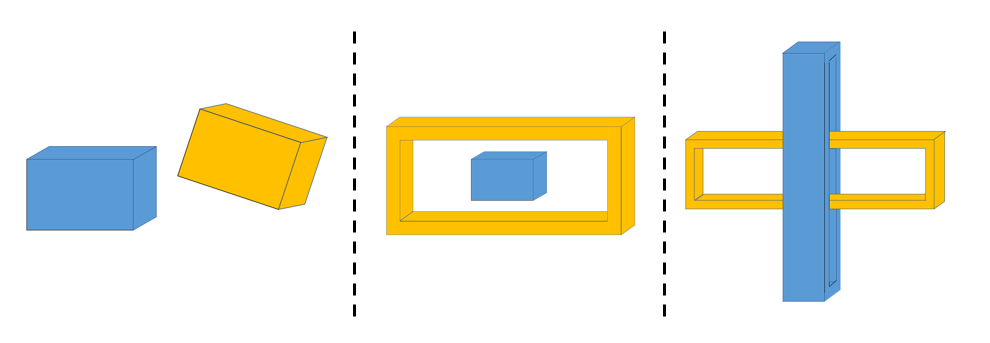}
    \caption{\label{fig:instance}Three scenes each with two objects. In all of them, the objects can be separated from each other based on their object features. Left: The objects have different locations and poses. Middle: The location and pose of each object are the same, but they differ in size. Right: The location and size of each object are the same, but they differ in pose.}
\end{figure}

To avoid the permutation-invariance problem in instance level segmentation, we use a new representation to indicate individual object instances. Let $^CP_k = \{(x_i,y_i,z_i)\}$ be a point cloud containing $N$ 3D points recorded from the camera frame $C$ while looking at $\mathcal{O}_k$. Let $^C{\bf \overline{x}}_k =(cx, cy, cz)^T$ denote the bounding box center of Object $\mathcal{O}_k$ with respect to the camera frame. Let $(\overline{xx}, \overline{yy}, \overline{zz}, \overline{xy}, \overline{yz}, \overline{zx})^T$ denote the second order moments of $\{^CP_k(x_i,y_i,z_i)-^C{\bf \overline{x}}_k\}$ where $\overline{xx} =\E((x_i - cx)(x_i - cx))$ and $\overline{xy} =\E((x_i - cx)(y_i - cy).$  As object feature, we define
\begin{align}
\xi_k =  ( & cx, cy, cz,  \nonumber \\ 
& cx+\overline{xx},cy+\overline{yy}, cz+\overline{zz}, \nonumber \\ 
& cx+\overline{xy}, cy+\overline{yz}, cz+\overline{zx}).
\end{align} 
Note that $(cx, cy, cz)$ encodes object location relative to camera frame $C$ while $(\overline{xx}, \overline{yy}, \overline{zz}, \overline{xy}, \overline{yz}, \overline{zx})^T$ encodes object size and pose. If a pixel $(u,v)$ in the input image shows object $\mathcal{O}_k$, then the correct output feature is $\xi_k$. We assume that different object instances will have different features with respect to the camera frame $C$. Therefore, pixels that show the same object will have the same object feature value while pixels belonging to different objects will differ in feature values. Three example scenes are visualized in Fig.~\ref{fig:instance}. We propose a neural network model to make per-pixel predictions of object features.

\subsection{Instance Segmentation Process}
The features $\hat{\xi}_{uv}$ predicted by our model are an approximation of the ground truth value. Each pixel (u,v) that corresponds to the same object $\mathcal O_k$ will have values $\hat{\xi}_{uv}$ that differ from ground truth by some $\epsilon_{uv}: \hat{\xi}_{uv} = \xi_{uv} + \epsilon_{uv}$. The predicted object feature values $\hat{\xi}_{uv}$ form a cluster around the ground truth value $\xi_{uv}$ under the error distribution $\epsilon_{uv}$. We propose to learn a model that provides the input to a clustering process in that feature space. Specifically, the model learns to predict each cluster's centroid and radius.

Inspired by region proposals~\citep{DBLP:journals/corr/RenHG015}, our model also outputs an image denoted by $\eta$. Each pixel $(u,v)$ in this image contains the probability $\eta_{uv}$ that it is the cluster centroid. To generate the ground truth of $\eta$, we sort pixels representing object $\mathcal{O}_k$ in the RGB images by their pixel distance to the object's centroid in ascending order. The top 10\% - 30\% of the ground truth object pixels per object in the input image will be annotated as cluster centroid candidates annotated as 1. The rest of the pixels are annotated as 0. 

Let $\mathcal{B}$ be an additional output image of our model. Pixel $(u,v)$ contains a scalar value $B_{uv}$. This value is a radius estimate of the sphere that encloses all pixels which belong to the same object. The sphere is centered at $\hat{\xi}_{uv}$. Any pixel at $(u,v)$ whose $\hat{\xi}_{uv}$ falls inside the sphere will be segmented as the same object $\mathcal{O}_k$. Any pixel at $(m,n)$ whose $\hat{\xi}_{mn}$ falls outside the sphere will be segmented as an object different from $\mathcal{O}_k$. To generate the ground truth $\mathcal{B}^{gt}$, each pixel (u,v) representing object $\mathcal{O}_k$ is annotated by half of the minimum distance between the object feature $\xi_k$ and $\xi_l~\forall l \neq k$ of all the other objects in the image:
\begin{equation}
B^{gt}_{uv} = \frac{1}{2} \min_{k \ne l}\|\xi_k - \xi_l\|_2
\end{equation}

\begin{figure}[t]
\centering
\includegraphics[width=0.9\columnwidth]{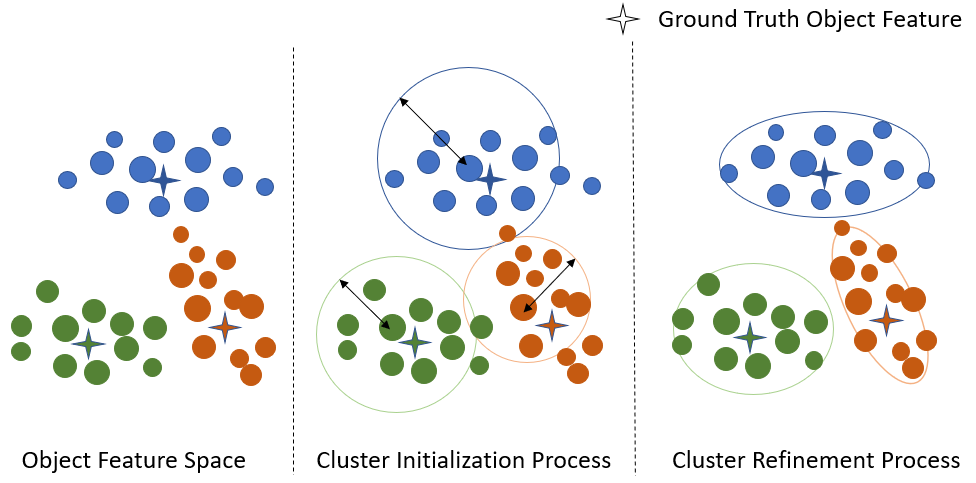}
\caption{Instance segmentation process. Left: Points represent predicted object features. Stars represent ground-truth object features. Same color indicates same object. The size of points represents the corresponding point's probability $\hat{\eta}$ of being an object centroid. Middle: Cluster initialization starts with points having the highest probability to be an object centroid. A sphere centered at one of those squares with radius $\hat{\mathcal{B}}$ then segments corresponding points. Right: After cluster initialization, we compute the mean and covariance of each cluster. These initialize a GMM. Refinement of instance segmentation is achieved by running the GMM one iteration.} 
\label{fig:cluster}
\end{figure}

Given the predicted $\hat{\mathcal{B}}$ and $\hat{\eta}$, we can now perform multi-object segmentation as visualized in Fig.~\ref{fig:cluster}. There are two stages of clustering. For initialization, pixel (u,v) with the maximum probability of being an object centroid $\hat{\eta}_{uv}$ is chosen first. Given a sphere centered at $\hat{\xi}_{uv}$ with radius $\hat{B}_{uv}$, all pixels $(m,n)$ with a feature $\hat{\xi}_{mn}$ enclosed by this sphere are assigned to object $O_1$. All pixel assigned to $O_1$ are removed from the set of unsegmented pixels before segmenting the next object. From the remaining pixels, the one with the highest $\hat{\eta}_{op}$ is used as the seed for segmenting $O_2$. This process is repeated until all foreground pixels are assigned to an object. After this initialization, we refine the resulting segmentation running one iteration of a {\em Gaussian Mixture Model\/} (GMM). The number of mixture components equals the number of object clusters. The covariance matrix of each mixture component equals the covariance matrix of the predicted object feature values in each cluster.

\subsection{ClusterNet Architecture}
\begin{figure*}
\centering
\includegraphics[width=0.95\linewidth]{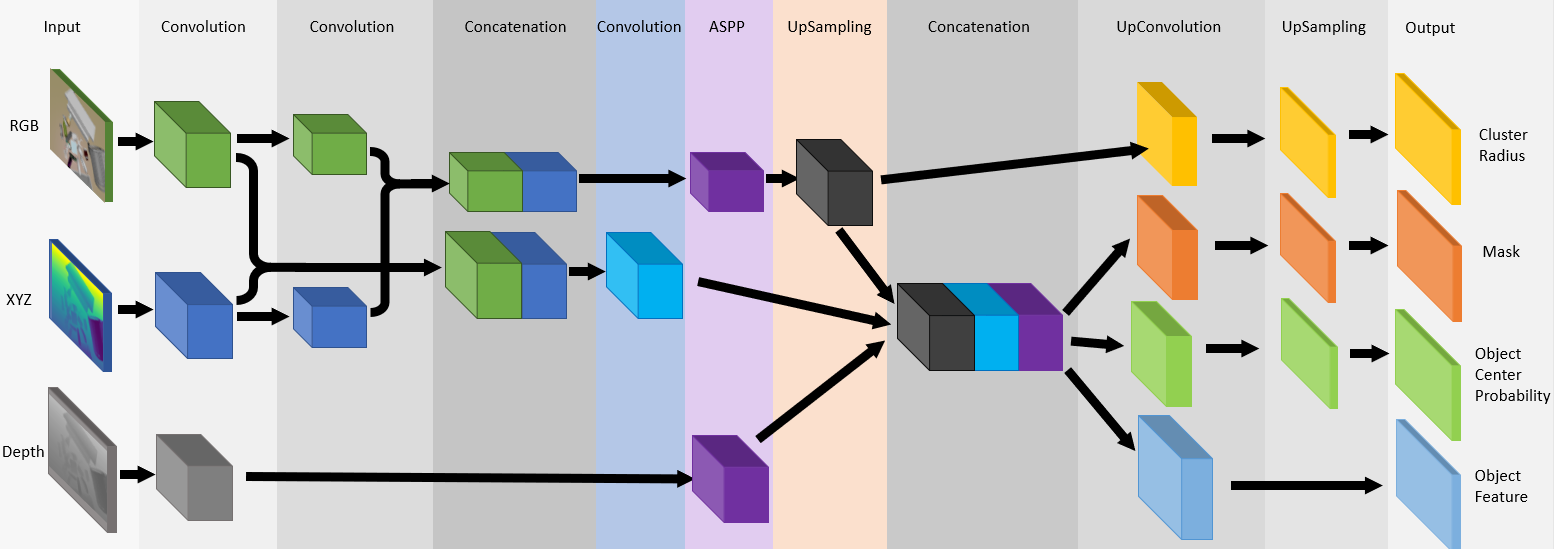}
\caption{Network architecture utilized in this paper. The input are three images: RGB, XYZ and Depth. The output of the upconvolution is fed into 4 different layers that make a per-pixel prediction of the object feature $\hat{\xi}$, the probability of the pixel being the object centroid $\hat{\eta}$, a foreground/background mask and the cluster radius $\hat{\mathcal{B}}$.} 
\label{fig:net}
\end{figure*}

Our proposed model is shown in Fig.~\ref{fig:net}. It takes three images as input: RGB, XYZ and depth. XYZ images are transformed from depth images using the camera intrinsics and therefore seem like redundant information. However, we will show in the experimental section that there is a gain in including both.
RGB and XYZ are fed into a pre-trained ResNet50~\cite{he2016deep} encoder. 

We extract their $C1$ level features with size of $(H/8,W/8,256)$ and $C4$ level features with size of $(H/32,W/32,2048)$ denoted as $C1_{rgb}$, $C1_{xyz}$ and $C4_{rgb}$, $C4_{xyz}$ respectively. Then $C1_{rgb}$ and $C1_{xyz}$ are concatenated and fed into another convolution layer to fuse them. The output is denoted as $C1_{f}$. $C4_{rgb}$ and $C4_{xyz}$ are concatenated to $C4_{rgbxyz}$ and fed into an {\em Atrous Spatial Pyramid Pooling\/} (ASPP) module \cite{chen2017rethinking} 
to enlarge the receptive field of each cell within $(W/32,H/32)$ of the feature map. The output is up-sampled to $(W/8,H/8)$ denoted as $C4_f$ to have the same size as $C1_f$. Depth images are fed into a VGG architecture~\cite{simonyan2014very} and fed into another ASPP module \cite{chen2017rethinking} denoted as $D_f$. Then $C1_f$, $C4_f$ and $D_f$ are concatenated as the final encoding stage $E_f$.  

Intuitively, the cluster radius could be inferred from global information like distances and poses between nearby objects. Therefore,  $\hat{\mathcal{B}}$ is directly decoded from $C4_f$. Cluster centroid probability images $\hat{\eta}$, mask images and object feature images are decoded from $E_f$. For cluster centroid probability images and mask images, the neural net model decodes them to $(W/4,H/4,2)$, $(W/4,H/4,2)$ respectively and then up-samples them by 4 using bilinear interpolation. For object feature images the neural network model decodes them directly to the original size $(W,H,9)$. 
\subsection{Loss function} \label{sec:loss}
\label{sec:loss}
We define the following  object-centric training loss:
\begin{align}
L =  \lambda_{s}L_{s}  +\lambda_{cen}L_{cen} + L_{p}  +\lambda_{var}L_{var}+\lambda_{vio}L_{vio}
\end{align}
where the $\lambda$s weight each loss term. Note that all pixel-wise loss terms $L_p$, $L_{cen}$, $L_{var}$ and $L_{vio}$ are computed only on the ground truth foreground pixels. 

\subsubsection{Semantic Mask Loss}
$L_{s}$ is the cross entropy loss between the ground truth and estimated semantic segmentation. In our experiment, there are only 2 classes (foreground/background). If a pixel is the projection of an object point, we assign 1 as ground truth; otherwise 0.

\subsubsection{Cluster Center Loss}
Cross-entropy loss $L_{cen}$ is used to learn the probability $\eta_{uv}$ of a pixel $(u,v)$ to be the object center.

\subsubsection{Pixel-wise Loss}
We use a pixel-wise loss $L_{p}$ on the object feature $\xi_{uv}$ and the enclosing sphere radius $B_{uv}$. For each attribute, we use the L2-norm to measure and minimize the error between predictions and ground truth. Note that the loss on each attribute is also differently weighted. We denote their corresponding weights $\lambda_{xi}$ and $\lambda_B$, respectively.

\subsubsection{Variance Loss}
We use $L_{var}$ to encourage pixels $(u,v)$ belonging to the same object $\mathcal{O}_{k}$ to have similar object features $\xi_{uv}$ and thereby to reduce their variance. 
\begin{equation}
L_{var} = \sum_{k}\frac{1}{N_{k}}\sum_{(u,v) \in \mathcal{O}_k}\| \hat{\xi}_{uv} - \bar{\hat{\xi}}_{uv} \|^2
\end{equation}

where $\bar{\hat{\xi}}_{uv}$ is the mean value of $\hat{\xi}_{uv}$ over all $N_k$ pixels belonging to $\mathcal{O}_k$. 

\subsubsection{Violation Loss}
$L_{vio}$ penalizes pixels $(u,v)$ that are not correctly segmented. Any predicted feature $\hat{\xi}_{uv}$ that is more than $\lambda_v\mathcal{B}_{uv}$ away from the ground truth $\xi_{uv}$ will be pushed towards the ground truth feature by the violation loss:

\begin{equation}
L_{vio} =\sum_{k}\sum_{(u,v)\in \mathcal{O}_k}  \mathbf{1}\{\|\hat{\xi}_{uv} -\xi_{uv}\|_2 > \lambda_v \mathcal{B}_{uv}\} \|\hat{\xi}_{uv} -\xi_{uv}\|_2 \nonumber \\
\end{equation}

where $\lambda_v$ is a hyperparameter to define the range of violation. In our experiments we found $\lambda_v=\frac{1}{5}$ to work well.

\section{Experiments}\label{sec:exp}
\subsection{Data Set}
For evaluation, we use the synthetic dataset by~\citet{Lin2018sceneflow}. It contains RGB-D images of scenes with a large variety of rigid objects. This dataset is very relevant to robotic manipulation research as it contains a wide variety of graspable objects and is recorded with an RGB-D camera that is very common on manipulation robots. Objects are also strongly occluding each other such that this dataset exposes limitations of current state-of-the-art approaches based on region-proposals. 

The dataset is generated from 31594 3D object mesh models from ShapeNet~\citep{shapenet2015} covering 28 categories. The models are split into a training, validation and test set with 21899, 3186  and 6509 objects respectively. For each scene, 1-30 object models are randomly selected. 49988, 6720, 14372 images are synthesized using models from training, validation and test sets respectively. 

\subsection{Evaluation Metrics}
We adopt the standard evaluation metrics like average precision (AP) and average recall (AR) as used for the COCO dataset~\citep{lin2014microsoft}. The average precision is measured based on intersection over union (IoU) between predicted segmentation and ground truth segmentation. $AP_{50}$ and $AP_{75}$ is the average precision based on an IoU threshold of $0.5$ and $0.75$ respectively. $AP$ is the average of $AP_x$ where x ranges from 0.5 to 0.95 in steps of $0.05$. The average recall is measured based on the maximum object segmentation candidates allowed. 
$AR_{1}$, $AR_{10}$ and $AR$ is calculated based on $1$, $10$ and $100$ maximum object segmentation candidates allowed. Objects are classified to be small, medium and large objects if their bounding box areas are within the range of $[0^2,32^2)$, $[32^2, 96^2)$ and $[96^2, 100000^2)$, respectively (units are pixels). AP and AR calculated on small, medium and large objects are denoted as $AP_{S}$, $AP_{M}$, $AP_{L}$ and $AR_{S}$, $AR_{M}$ and $AR_{L}$. We also define an object occlusion score. It constitutes the number of pixels showing the occluded object divided by the number of object pixels if the object was not occluded. Objects are classified to be under heavy, medium and little occlusion if their occlusion score is within the range of $[0,0.3)$, $[0.3, 0.75)$ and $[0.75, 1.0]$, respectively. AR calculated on objects under heavy, medium and little occlusion is denoted as $AR_{\text{HO}}$, $AR_{\text{MO}}$ and $AR_{\text{LO}}$.

\subsection{Baselines}
We compare our method with Mask R-CNN~\cite{he2017mask} that only uses RGB images as input. As backbone, we choose ResNet-50-C4 denoted as \textbf{Mask R-CNN(C4)} and ResNet-50-FPN denoted as \textbf{Mask R-CNN(FPN)}. We use the Detectron~\cite{Detectron2018} implementation and its default parameters. The experiment is running on two Nvidia P100 GPUs with batch size 2 per GPU for 200000 iterations. We change the number of classes to be 2 only containing \textbf{Foreground} and \textbf{Background} and use RLE format~\citep{lin2014microsoft} as the ground truth annotation format. 

\begin{table*}[ht!]
\renewcommand\arraystretch{1}
\centering
\resizebox{\textwidth}{!}{
\begin{tabular}{c | c c c c c c c c c c c c}
\hline
\hline 
 method & $\text{AP}$ & $\text{AP}_{50}$ & $\text{AP}_{75}$ & $\text{AP}_{S}$ & $\text{AP}_{M}$ &  $\text{AP}_{L}$ & $\text{AR}$ & $\text{AR}_{1}$ & $\text{AR}_{10}$ & $\text{AR}_{S}$ & $\text{AR}_{M}$ & $\text{AR}_{L}$\\
\hline
Mask R-CNN(C4) &53.1 &82.1 &56.1 & 8.0&50.2 & 63.5 & 61.3& 8.4 &52.2 & 28.3& 61.0 & 68.7\\
Mask R-CNN(FPN) &51.4 &84.0 &54.9 & 9.1 &47.8 &61.5 & 59.9& 8.3&51.2& 31.3 & 59.2 & 66.9\\
\hline 
Our(rgb$\to$c) & 24.9 & 46.1 & 23.5 & 7.0 & 26.0 & 28.6 & 41.5 &5.8 & 35.1 &  10.7 & 38.2 & 52.1\\
Our(rgbd$\to$c) & 36.4 & 60.2 & 36.1 & 15.5 & 39.9& 36.6& 52.4&7.6 & 43.3 & 22.2 & 52.7 & 58.5\\
Our(rgbxyz$\to$c)& 63.7&  83.0& 68.2 & 24.6 & 63.6 & 72.7 & 70.3 & 9.7 & 60.3& \textbf{31.4}& 69.4 & 79.6\\
Our(c) & \textbf{66.2} & \textbf{86.4} & \textbf{71.6} & \textbf{25.2} & \textbf{64.9} & \textbf{76.3} & \textbf{71.3} & \textbf{10.0} &\textbf{61.8} & 30.8 & \textbf{69.8} & \textbf{81.9} \\
Our(c+cov) & 60.9 & 82.8 & 66.2 & 23.1 & 61.4 & 68.8 & 69.1 & 8.6 & 59.0 & 28.0 & 67.7 &79.7\\
\hline
\hline
\end{tabular}}
\vspace{0.5em}
\caption{Performance of instance segmentation on our synthesized dataset.}
\label{tab::exp}
\end{table*}

\begin{table}[ht!]
  \centering
  \centering 
 \begin{tabular}{c | c c c}
\hline
\hline 
Method  & $\text{AR}_{\text{HO}}$ & $\text{AR}_{\text{MO}}$ & $\text{AR}_{\text{LO}}$ \\ 
\hline
Mask R-CNN(C4) & 41.5 & 56.7 & 69.2 \\
Mask R-CNN(FPN) & 41.9 & 54.8& 67.5 \\
\hline
Our(rgbxyz$\to$c)& 49.6 &66.2 & 78.3\\
Our(c) &  \textbf{50.7} & \textbf{68.5} & \textbf{78.8}\\
Our(c+cov) &47.6 &66.6 &76.7\\
\hline
\hline 
\end{tabular}
\caption{\label{tab:expB}Average recall results under different occlusion conditions.}
\end{table}

\subsection{Quantitative Evaluations of Instance Segmentation}
We refer to the network in Fig.~\ref{fig:net} as
\textbf{Our(c+cov)} where \textbf{c+cov} represents using both object center and object covariance matrix as object feature $\xi$. We also compare to a variant of our proposed architecture $\textbf{Our(c)}$ where the object feature $\xi$ only contains its 3D centroid. This provides insight into the impact of using the second order moments as object features. Furthermore, we run three ablation studies to demonstrate the impact of different input image modalities. We denote \textbf{Our(rgbd$\to$c)} as neural network that only uses RGB and depth images as the input. \textbf{Our(rgb$\to$c)} denotes the model that use only RGB images as input. \textbf{Our(rgbxyz$\to$c)} use both RGB and point clouds as inputs. \textbf{Our(rgbd$\to$c)}, \textbf{Our(rgb$\to$c)} and \textbf{Our(rgbxyz$\to$c)} all use only the object centroid as object feature $\xi$.

\begin{figure*}[ht!]
  \centering
  \begin{tabular}{@{}|@{}c@{}|@{}|@{}c@{}|@{}|@{}c@{}|@{}c@{}|@{}|@{}c@{}|@{}|@{}c@{}|@{}|@{}c@{}|@{}c@{}|}
    \hline 
    \small RGB & \small GT & Our & \small MR  & \small RGB  & \small GT & \small Our & \small MR\\
   \hline
	\begin{minipage}{0.125\linewidth}
  \centering 
      \includegraphics[width=\textwidth]{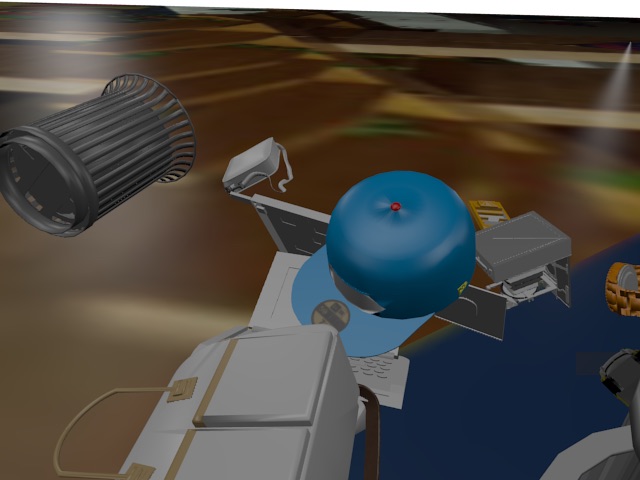}
    \end{minipage}
    &  
 \begin{minipage}{0.125\linewidth}
 \centering
   \includegraphics[width=\textwidth]{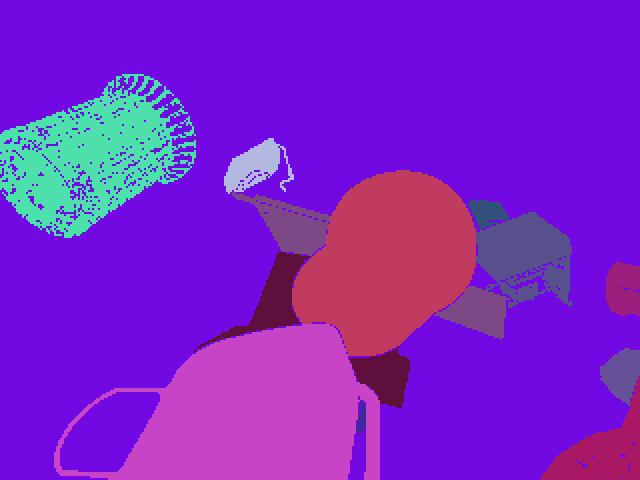}
   \end{minipage}%
    & 
  \begin{minipage}{0.125\linewidth}
  \centering 
      \includegraphics[width=\textwidth]{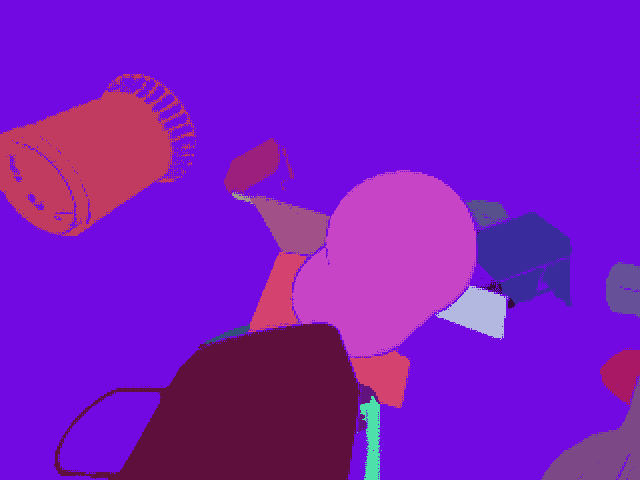}
   \end{minipage}
    & 
 \begin{minipage}{0.125\linewidth}
    \centering 
      \includegraphics[width=\textwidth]{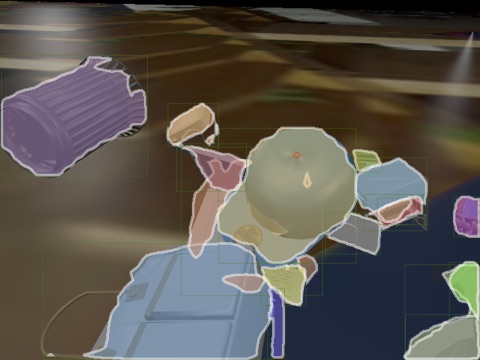}
    \end{minipage}
       &
	\begin{minipage}{0.125\linewidth}
  \centering 
      \includegraphics[width=\textwidth]{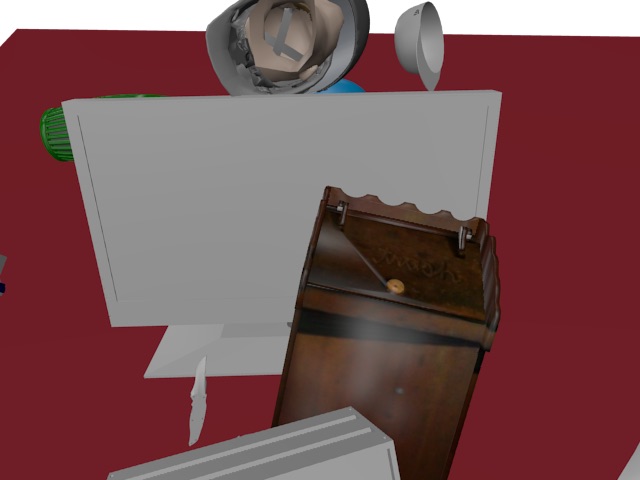}
    \end{minipage}
    &   
 	\begin{minipage}{0.125\linewidth}
 \centering
   \includegraphics[width=\textwidth]{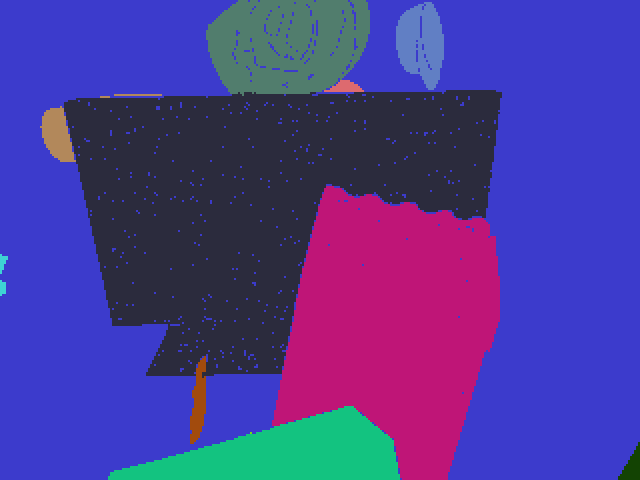}
   \end{minipage}
   & 
  \begin{minipage}{0.125\linewidth}
  \centering 
      \includegraphics[width=\textwidth]{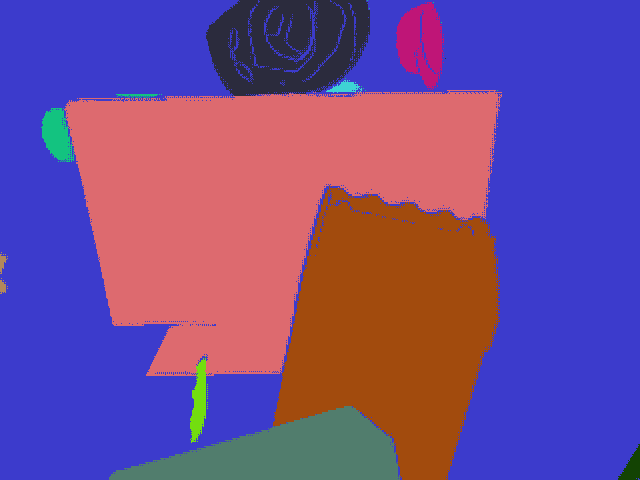}
   \end{minipage}
    & 
 	\begin{minipage}{0.125\linewidth}
      \includegraphics[width=\textwidth]{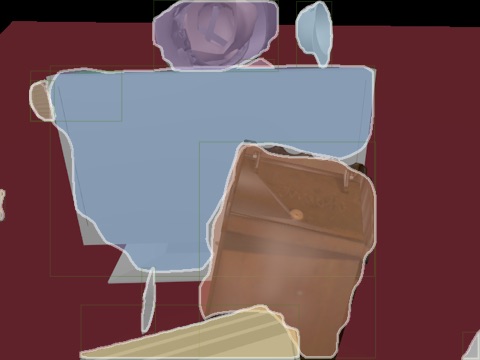}
    \end{minipage}\\
           \hline
	\begin{minipage}{0.125\linewidth}
  \centering 
      \includegraphics[width=\textwidth]{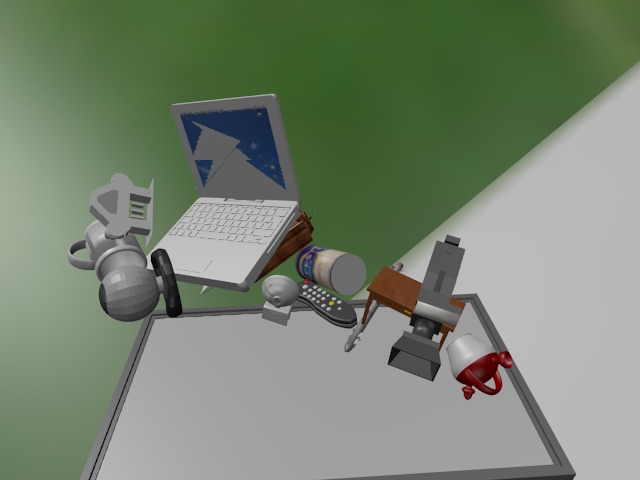}
    \end{minipage}
    &   
 	\begin{minipage}{0.125\linewidth}
 \centering
      \includegraphics[width=\textwidth]{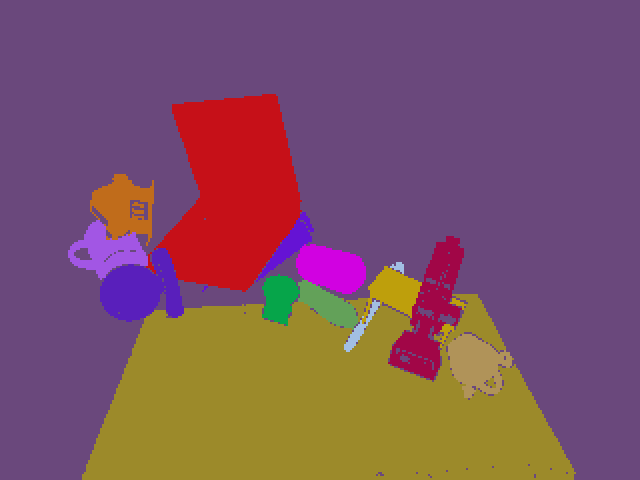}
   \end{minipage}
    & 
  \begin{minipage}{0.125\linewidth}
  \centering 
      \includegraphics[width=\textwidth]{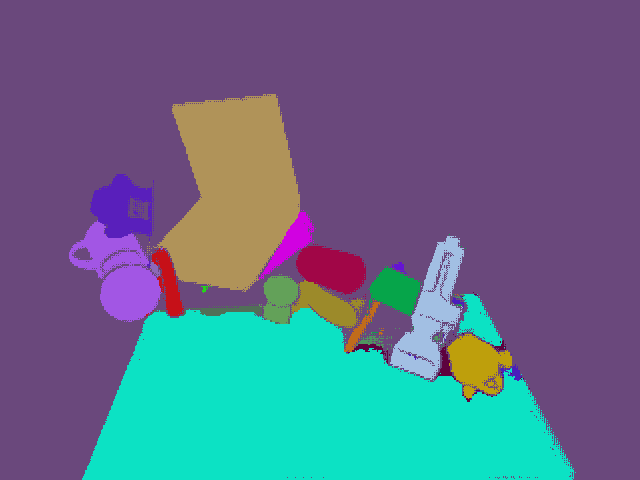}
   \end{minipage}
    & 
 	\begin{minipage}{0.125\linewidth}
      \includegraphics[width=\textwidth]{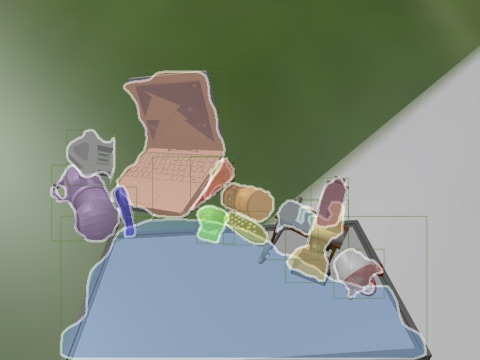}
    \end{minipage}
      &
	\begin{minipage}{0.125\linewidth}
  \centering 
      \includegraphics[width=\textwidth]{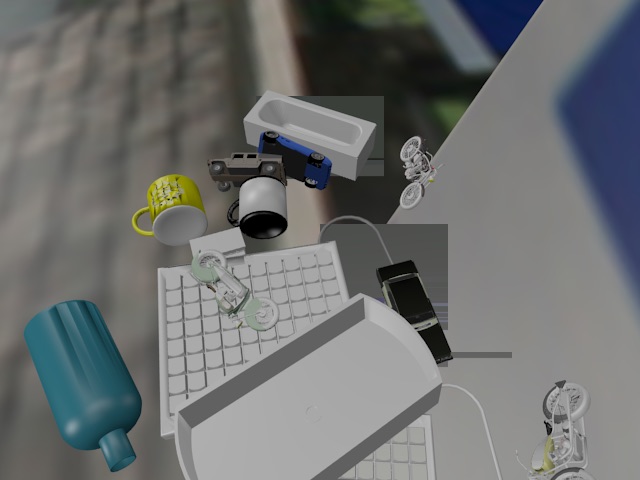}
    \end{minipage}
    &   
 	\begin{minipage}{0.125\linewidth}
 \centering
      \includegraphics[width=\textwidth]{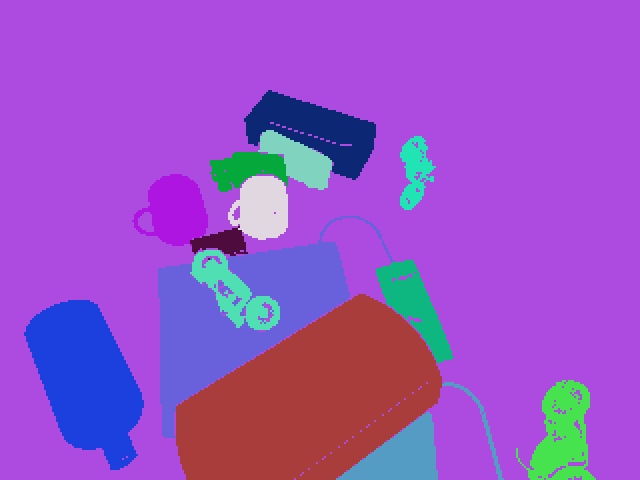}
   \end{minipage}
    & 
  \begin{minipage}{0.125\linewidth}
  \centering 
      \includegraphics[width=\textwidth]{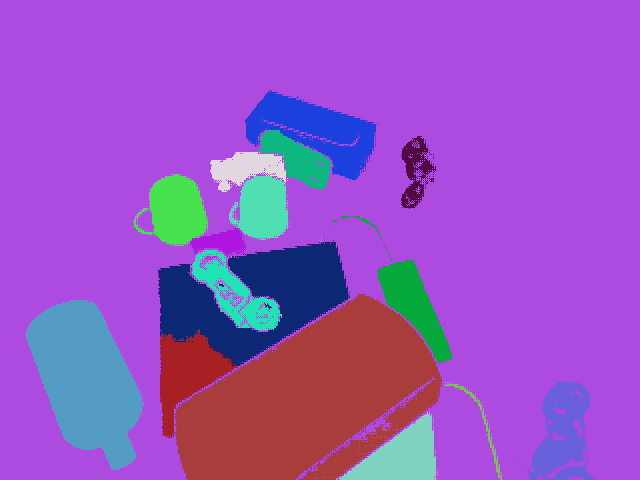}
   \end{minipage}
    & 
 	\begin{minipage}{0.125\linewidth}
      \includegraphics[width=\textwidth]{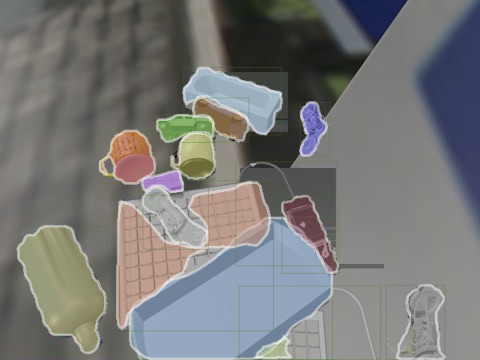}
    \end{minipage}\\
               \hline
	\begin{minipage}{0.125\linewidth}
  \centering 
      \includegraphics[width=\textwidth]{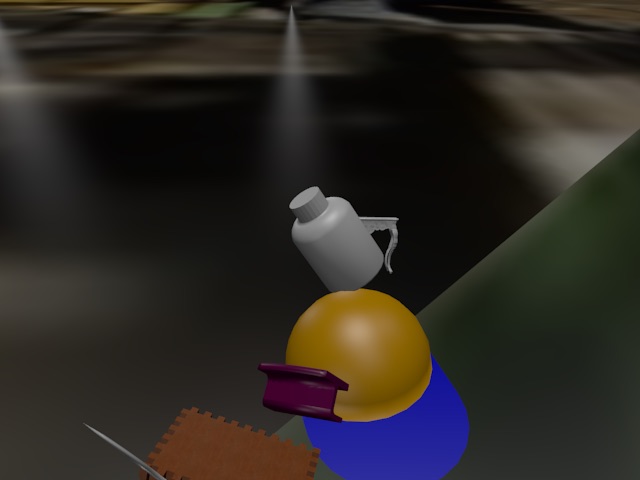}
    \end{minipage}
    &   
 	\begin{minipage}{0.125\linewidth}
 \centering
      \includegraphics[width=\textwidth]{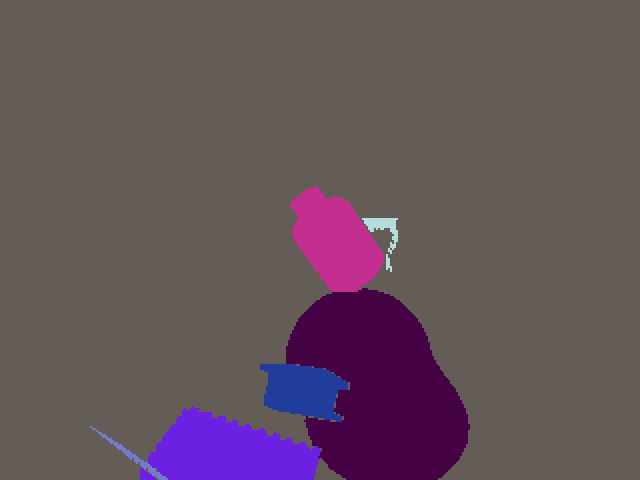}
   \end{minipage}
    & 
  \begin{minipage}{0.125\linewidth}
  \centering 
      \includegraphics[width=\textwidth]{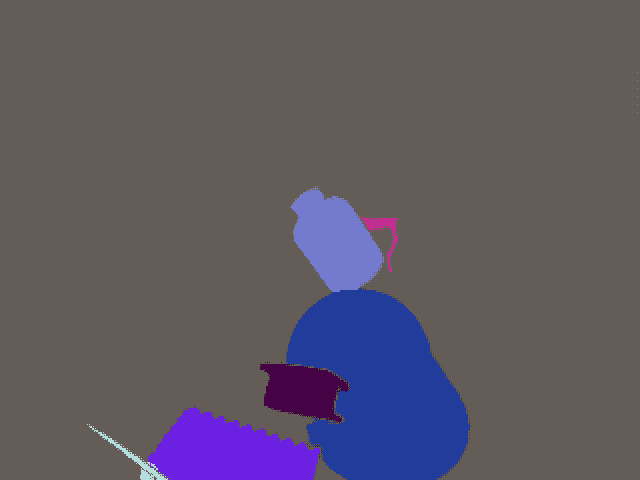}
   \end{minipage}
    & 
 	\begin{minipage}{0.125\linewidth}
      \includegraphics[width=\textwidth]{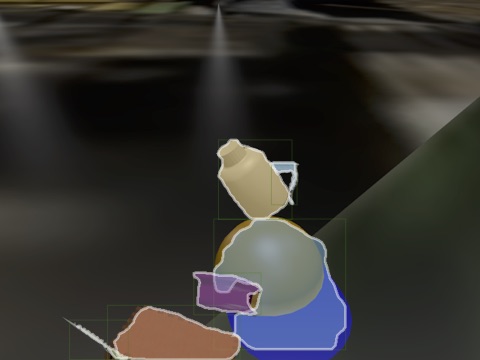}
    \end{minipage}
      &
	\begin{minipage}{0.125\linewidth}
  \centering 
      \includegraphics[width=\textwidth]{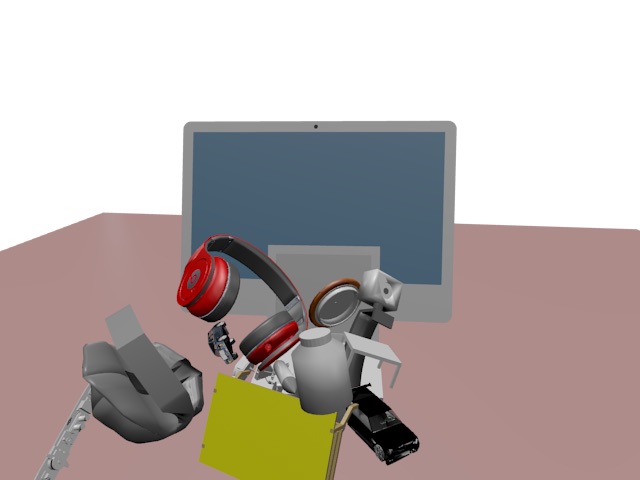}
    \end{minipage}
    &   
 	\begin{minipage}{0.125\linewidth}
 \centering
      \includegraphics[width=\textwidth]{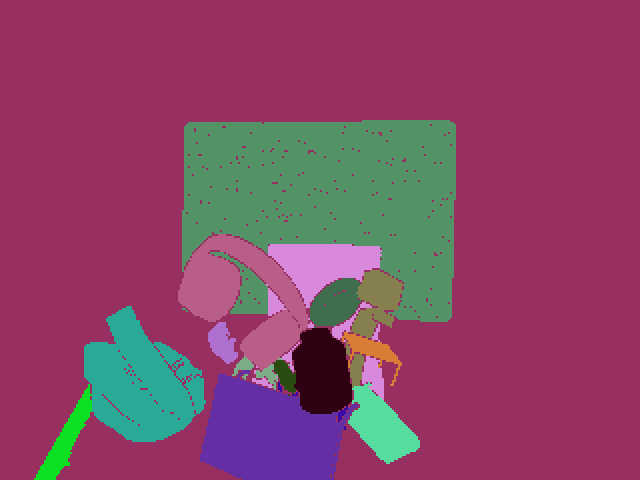}
   \end{minipage}
    & 
  \begin{minipage}{0.125\linewidth}
  \centering 
      \includegraphics[width=\textwidth]{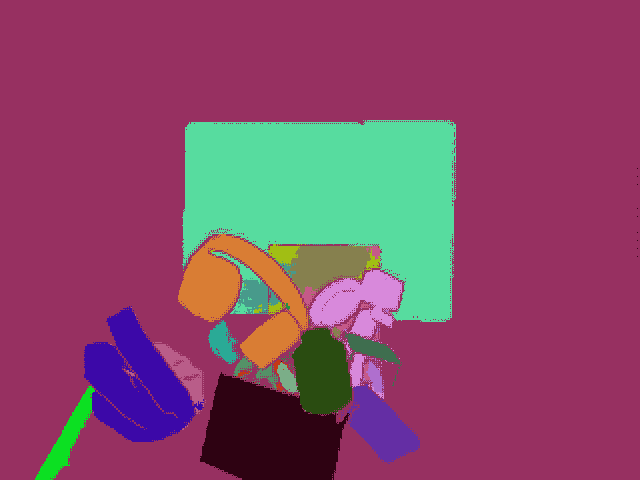}
   \end{minipage}
    & 
 	\begin{minipage}{0.125\linewidth}
      \includegraphics[width=\textwidth]{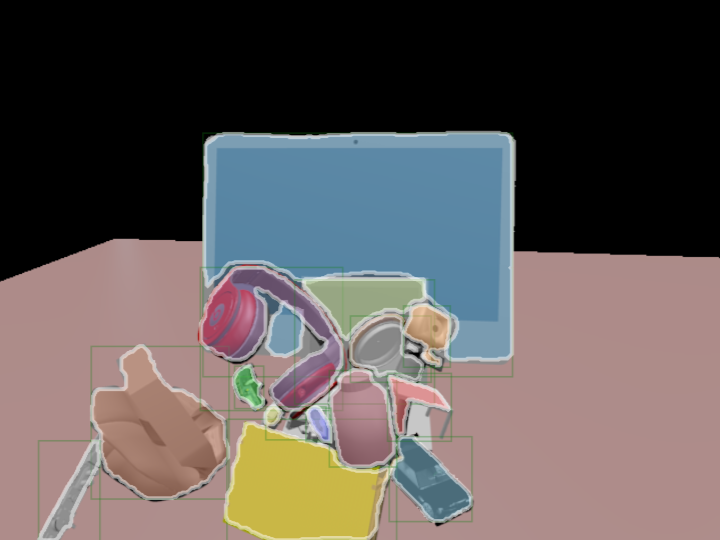}
    \end{minipage}\\
  \end{tabular}
\caption{Performance comparison of the proposed method on our synthetic dataset. RGB columns show RGB inputs. GT columns represent ground truth segmentation annotations. Our columns correspond to segmentation results using model \textbf{Our(c)}. MR columns correspond to segmentation results using  \textbf{Mask R-CNN(C4)}}\label{fig:qual_comp}
\end{figure*}

For training, we use the Adam optimizer~\citep{kingma2014adam} with its
suggested default parameters of $\beta1=0.9$ and
$\beta2=0.999$ along with a learning rate $\alpha=0.0001$~\cite{kingma2014adam}. We use a batch size of 4 image pairs. The input RGB-D images have a resolution of $480\times 640$. The loss weights, as defined in Sec.~\ref{sec:loss}, are set to $\lambda_m=1.0$, $\lambda_{cen}=1.0$, $\lambda_{var}=1$,
$\lambda_{vio}=1$, $\lambda_{p}$=100.0, $\lambda_{xi}=1.0$ and $\lambda_{B}=10.0$. The predicted object features $\hat{\xi}$ becomes more accurate after epoch 5. Then we set $\lambda_{var}=100.0$ and $\lambda_{vio}=100.0$ to increase the impact of the variance and violation loss. 
The results in terms of instance segmentation accuracy are shown in Table~\ref{tab::exp}. Our proposed models \textbf{Ours(c)} and \textbf{Ours(c+cov)} outperform the aforementioned Mask R-CNN~\cite{he2017mask} approaches by a large margin. The results in terms of segmentation accuracy under different levels of occlusions are reported in Table \ref{tab:expB}. A few example results are shown in Fig.~\ref{fig:qual_comp}. Compared to the Mask R-CNN~\cite{he2017mask} baselines, our models improve the average recall of objects under all levels of occlusions. 

\subsubsection{Effects of using different input image modalities}
We report the instance segmentation performance in Table~\ref{tab::exp}. $\textbf{Our(rgb$\to$c)}$ has the lowest performance of all models including the Mask R-CNN~\cite{he2017mask} baselines that only take RGB images as input. This is because the 3D centroid of an object is hard to reconstruct from RGB images. Including depth data ($\textbf{Our(rgbd$\to$c)}$) already improves the performance by a large margin compared with $\textbf{Our(rgb$\to$c)}$.  Including point cloud information ($\textbf{Our(rgbxyz$\to$c)}$) improves the instance segmentation performance further and outperforms MaskR-CNN significantly. This indicates that point clouds provide strong cues about 3D object instances and that our model leverages these cues well. $\textbf{Our(c)}$ relies on both depth images and point clouds. Compared with $\textbf{Our(rgbxyz$\to$c)}$, $\textbf{Our(c)}$ achieves improved results especially on large objects ($AP_L$ and $AR_L$). It demonstrates that depth images provide useful cues to segment large objects. This is probably due to their large variance in the X,Y channel of the point clouds.

\subsubsection{Effects of using different object features $\xi$}
$\textbf{Our(c)}$ shows better performance than $\textbf{Our(c+cov)}$. Predicting second order moments is more difficult than predicting the object centroid. This is suggested by the high variance of errors when predicting this feature. Additionally, there are only few objects in the dataset that expose situations as shown in the middle and right of Fig~\ref{fig:instance}. 

\subsection{Qualitative Evaluation on Real World Data}
We demonstrate the network's ability to perform instance segmentation in a real world setting. We recorded real RGB-D data with the Intel RealSense SR300 Camera. It was captured using a diverse set of objects with varying geometries, textures, and colors.  Note that we do not have any ground truth annotations. 

\begin{figure*}[ht!]
  \centering
  \begin{tabular}{@{}|@{}c@{}|@{}|@{}c@{}|@{}|@{}c@{}|@{}c@{}|@{}|@{}c@{}|@{}|@{}c@{}|@{}|}
    \hline 
    \small RGB  & MR & \small Our & \small RGB  & \small MR & \small Our\\
   \hline
  \begin{minipage}{0.125\linewidth}
  \centering 
      \includegraphics[width=\textwidth]{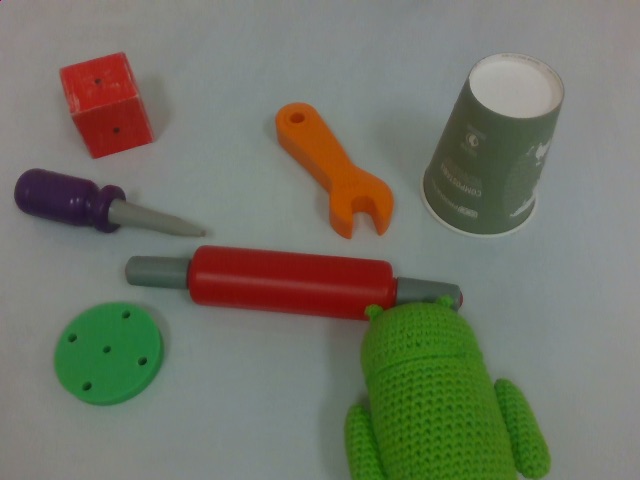}
   \end{minipage}
    & 
 \begin{minipage}{0.125\linewidth}
    \centering 
      \includegraphics[width=\textwidth]{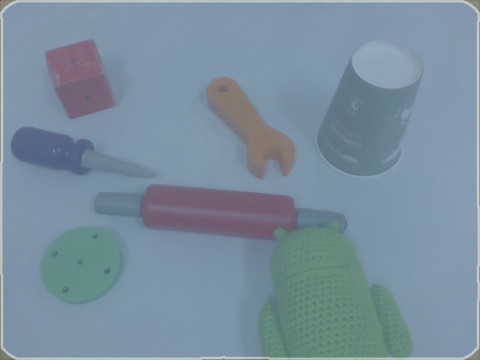}
    \end{minipage}
       &
	\begin{minipage}{0.125\linewidth}
  \centering 
      \includegraphics[width=\textwidth]{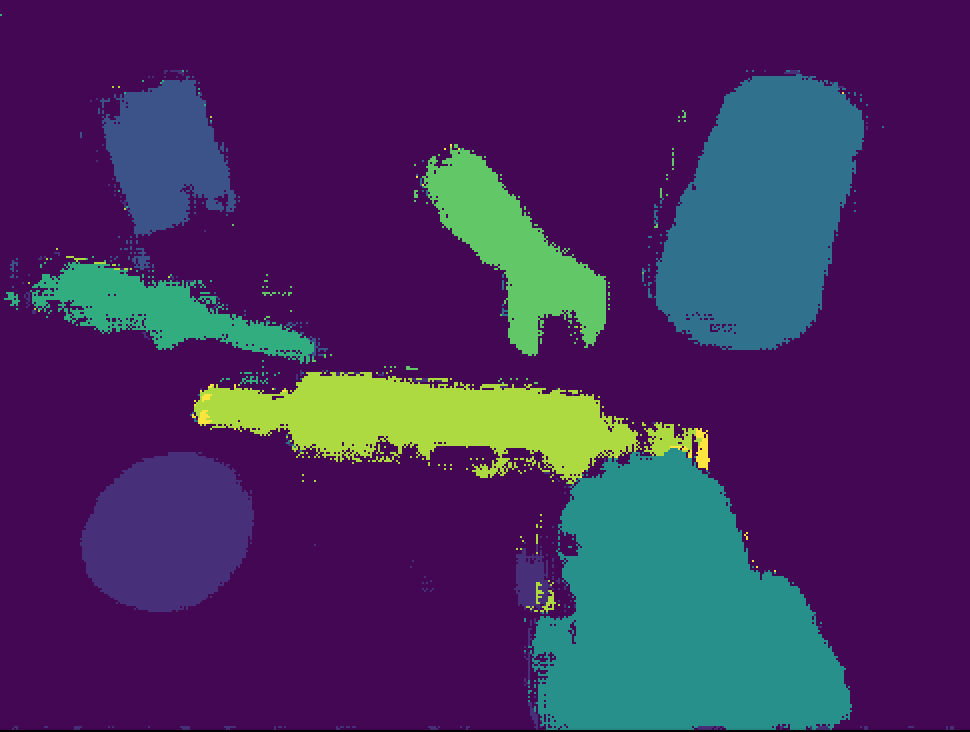}
    \end{minipage}
    &   
 	\begin{minipage}{0.125\linewidth}
 \centering
   \includegraphics[width=\textwidth]{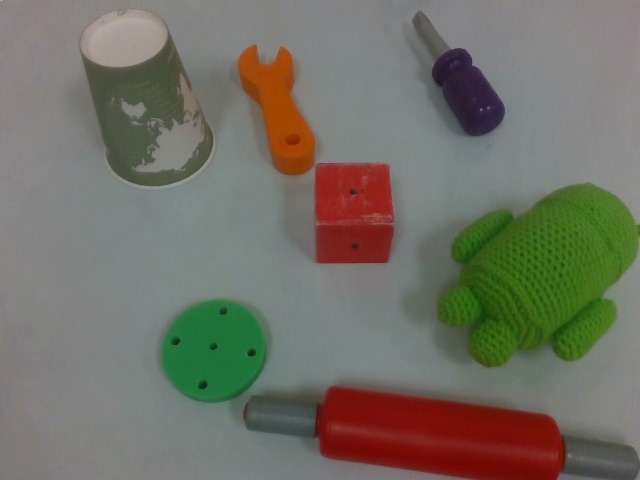}
   \end{minipage}
   & 
  \begin{minipage}{0.125\linewidth}
  \centering 
      \includegraphics[width=\textwidth]{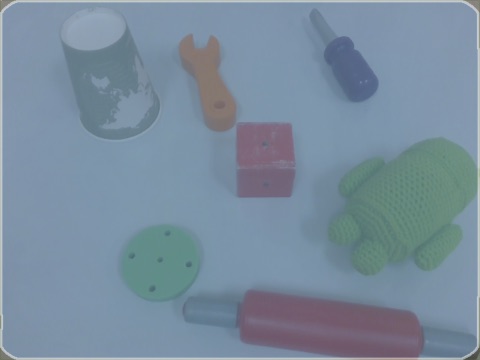}
   \end{minipage}
    & 
 	\begin{minipage}{0.125\linewidth}
      \includegraphics[width=\textwidth]{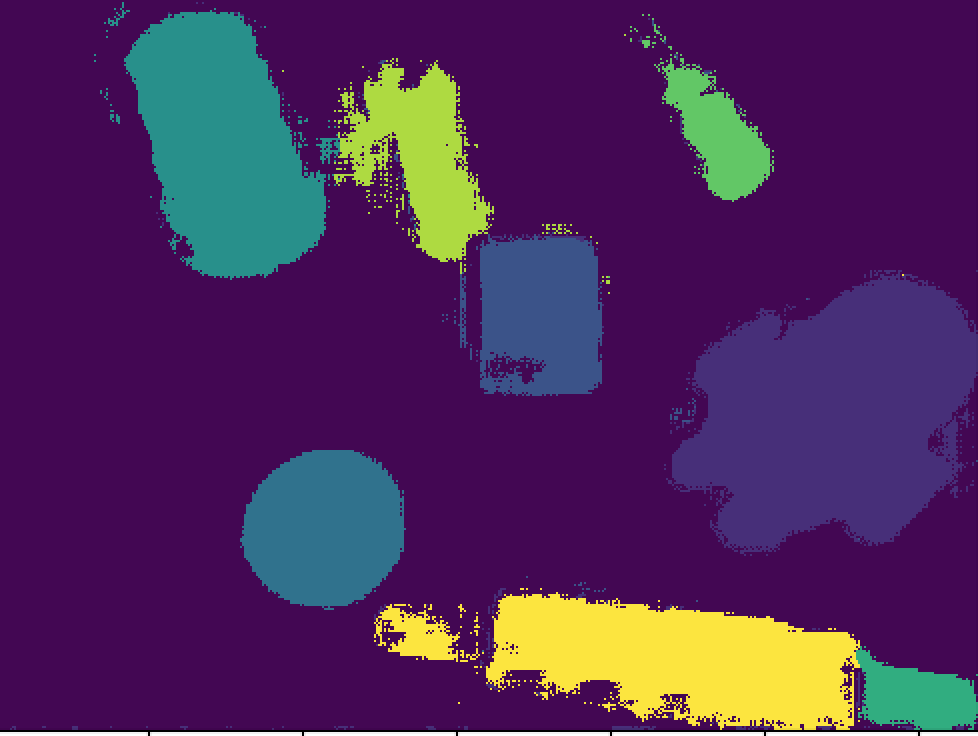}
    \end{minipage}\\
           \hline
  \begin{minipage}{0.125\linewidth}
  \centering 
      \includegraphics[width=\textwidth]{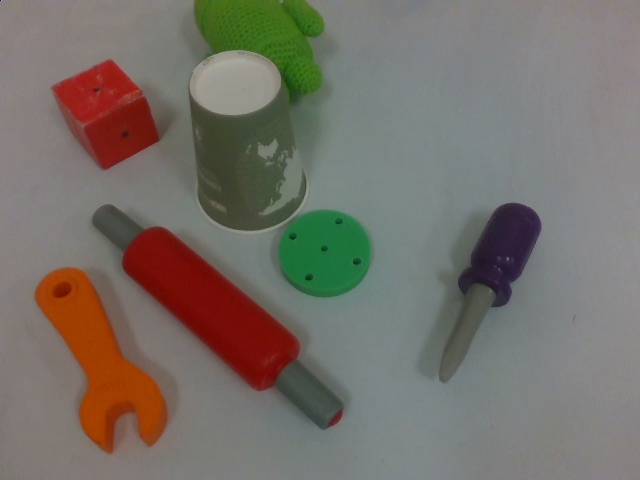}
   \end{minipage}
      &
	\begin{minipage}{0.125\linewidth}
  \centering 
      \includegraphics[width=\textwidth]{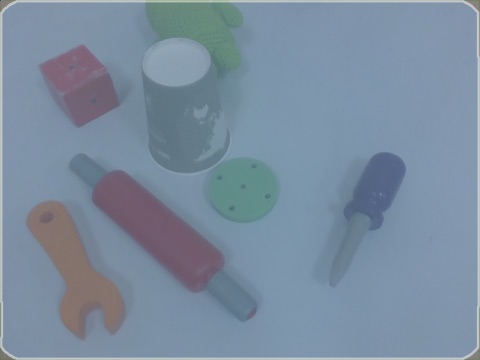}
    \end{minipage}
    &   
 	\begin{minipage}{0.125\linewidth}
 \centering
      \includegraphics[width=\textwidth]{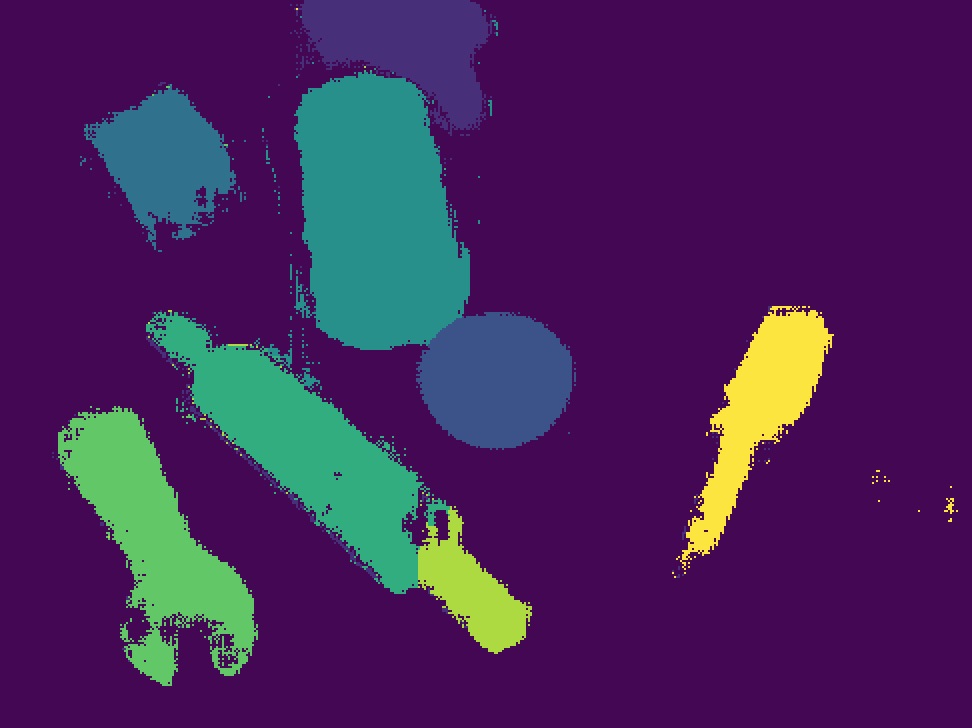}
   \end{minipage}
    & 
  \begin{minipage}{0.125\linewidth}
  \centering 
      \includegraphics[width=\textwidth]{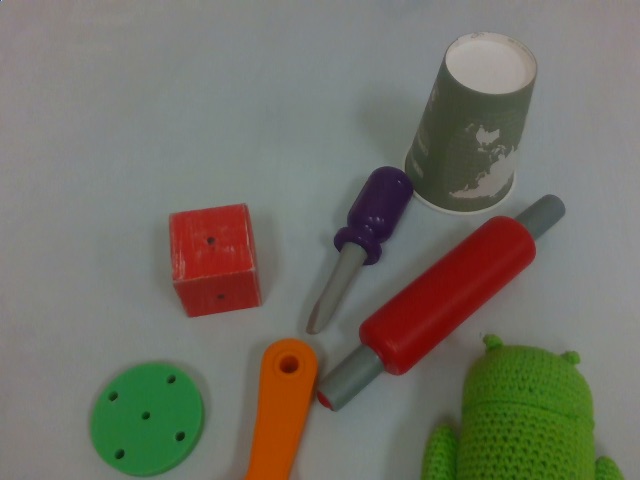}
   \end{minipage}
    & 
 	\begin{minipage}{0.125\linewidth}
      \includegraphics[width=\textwidth]{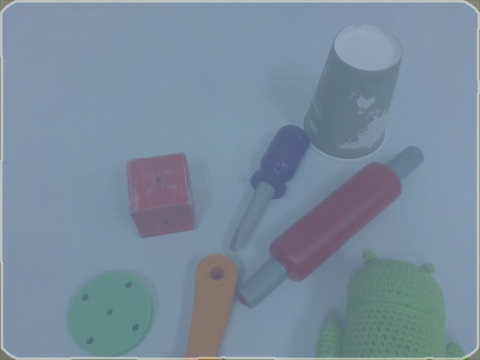}
    \end{minipage}
    & 
 	\begin{minipage}{0.125\linewidth}
      \includegraphics[width=\textwidth]{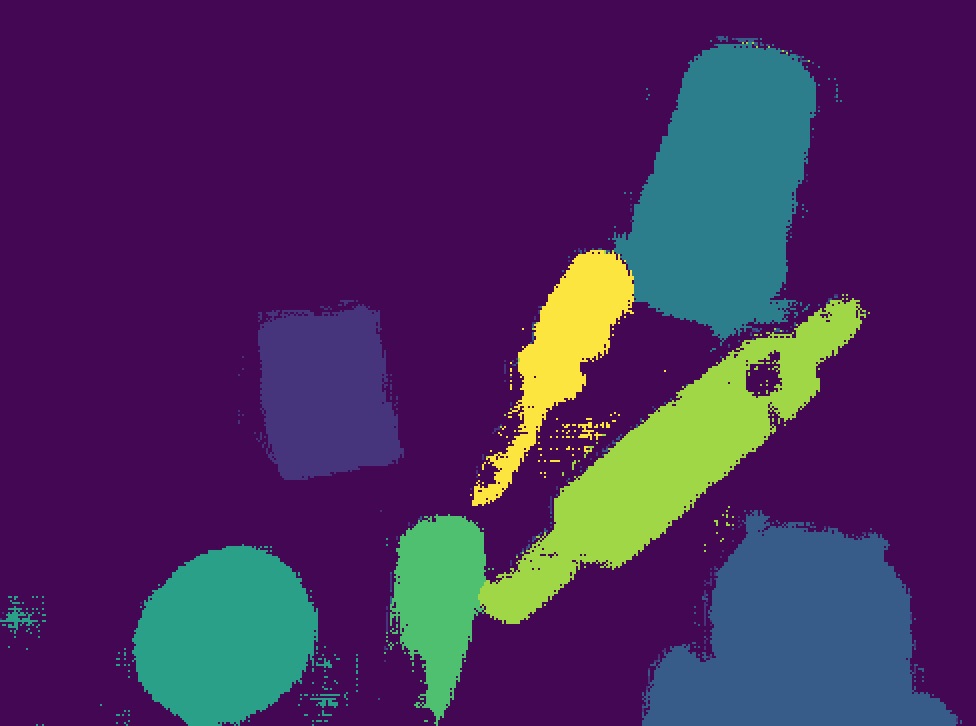}
    \end{minipage}\\
  \end{tabular}
  \caption{Performance comparison of the proposed method on the real world data. {\em RGB\/} columns show RGB input image. {\em MR\/} columns show the result of \textbf{Mask R-CNN(C4)}. {\em Our\/} columns correspond to segmentation results using \textbf{Our(c)} model. }\label{fig:real_comp}
\end{figure*}

We compare the performance of \textbf{Mask R-CNN(C4)} \citep{he2017mask} and our method \textbf{Our(c)}. Note that neither model is fine-tuned to transfer from synthetic to real data. The qualitative results are shown in Fig.~\ref{fig:real_comp}. Our method generalizes well and is able to accurately segment object instances. However \textbf{Mask R-CNN(C4)} segments the whole image as an instance and fails to detect and segment any objects indicating poor generalization ability.

\section{Conclusion}\label{sec:conclusion}
We proposed ClusterNet, a model towards 3D instance segmentation of an unknown number of objects in RGB-D images. In our approach, instance segmentation is formulated as a regression problem towards 3D object features that is followed by clustering.
Specifically, the model makes pixel-wise predictions of the first and second order moment of the object that the pixel belongs to. Then, sequential clustering is performed in this feature space to infer the object instances. Through this formulation, we showed how our model can leverage RGB {\em and\/} depth data to achieve robust 3D instance segmentation in challenging, cluttered scenes with heavy occlusions. While these quantitative results were on synthetic data, we also showed how our model transfers to similarly challenging real world data. As future work, we aim to evaluate this model also on datasets for autonomous driving such as Cityscapes~\citep{2016Cityscape} or ApolloScape~\citep{ApolloScape}. We would also like to integrate this approach with semantic segmentation  that is currently restricted to foreground/background segmentation.

\bibliographystyle{plainnat}
\bibliography{example} 

\end{document}